\documentclass[11pt,a4paper]{article}
\usepackage[hyperref]{acl2020}
\usepackage{times}
\usepackage{latexsym}
\usepackage{siunitx}
\usepackage{amsmath}
\usepackage{url}
\usepackage{multirow}
\usepackage{graphicx}
\usepackage{graphics}
\usepackage{enumitem}
\usepackage{subcaption}
\usepackage{caption}
\usepackage{amssymb}
\usepackage{verbatim}
\usepackage[normalem]{ulem} 
\usepackage{makecell}
\usepackage{array}
\usepackage{url}
\usepackage{comment}
\usepackage{multirow}
\usepackage{gensymb}
\usepackage[symbol]{footmisc}

\def\startcite#1{\citeauthor{#1} (\citeyear{#1})}
\usepackage{microtype}

\aclfinalcopy 

\title{Towards Transparent and Explainable Attention Models}

\author{Akash Kumar Mohankumar$^{1}$ \hspace{0.2cm}  Preksha Nema$^{1,2}$ \hspace{0.2cm} Sharan Narasimhan$^{1}$ \\ \bf{Mitesh M.  Khapra}$^{1,2}$ \hspace{0.1cm} {Balaji Vasan Srinivasan}$^{3}$ \hspace{0.1cm} Balaraman Ravindran$^{1,2}$ \\ \\
  $^1$Indian Institute of Technology Madras \hspace{0.1cm} 
  $^{3}$Adobe Research\\
   $^{2}$ Robert Bosch Center for Data Science and Artificial Intelligence, IIT Madras \\
  \texttt{\small makashkumar99@gmail.com} \hspace{0.1cm} \texttt{\small {\{preksha,miteshk,ravi\}@cse.iitm.ac.in}} \\ \texttt{\small sharan.n21@gmail.com} \hspace{0.1cm} \texttt{\small balsrini@adobe.com } 
}
\date{}

\begin{document}
\maketitle

\begin{abstract}
Recent studies on interpretability of attention distributions have led to notions of \textit{faithful} and \textit{plausible} explanations for a model's predictions. Attention distributions can be considered a faithful explanation if a higher attention weight implies a greater impact on the model's prediction. They can be considered a plausible explanation if they provide a human-understandable justification for the model's predictions. In this work, we first explain why current attention mechanisms in LSTM based encoders can neither provide a faithful nor a plausible explanation of the model's predictions. We observe that in LSTM based encoders the hidden representations at different time-steps are very similar to each other (high conicity) and attention weights in these situations do not carry much meaning because even a \textit{random} permutation of the attention weights does not affect the model's predictions. Based on experiments on a wide variety of tasks and datasets, we observe attention distributions often attribute the model's predictions to unimportant words such as punctuation and fail to offer a plausible explanation for the predictions. To make attention mechanisms more faithful and plausible, we propose a modified LSTM cell with a diversity-driven training objective that ensures that the hidden representations learned at different time steps are diverse. We show that the resulting attention distributions offer more transparency as they (i) provide a more precise importance ranking of the hidden states (ii) are better indicative of words important for the model's predictions (iii) correlate {better} with gradient-based attribution methods. Human evaluations indicate that the attention distributions learned by our model offer a plausible explanation of the model's predictions. Our code has been made publicly available at \url{https://github.com/akashkm99/Interpretable-Attention}
\end{abstract}
\section{Introduction}

\begin{table}[h]
    \small
    \begin{tabular}{p{7cm}}
        \Xhline{3\arrayrulewidth}
        \textbf{Question 1:} What is the best way to improve my spoken English soon ?\\
        \textbf{Question 2:} How can I improve my English speaking ability ?\\
        \textbf{Is paraphrase} (Actual \& Predicted): Yes \\
         \hline 
         {Attention Distribution} \\
            \begin{tabular}{l|l}
                {Vanilla LSTM}& 
                {\setlength{\fboxsep}{0pt}\colorbox[Hsb]{202, 0.00, 1.0}{\strut How}} {\setlength{\fboxsep}{0pt}\colorbox[Hsb]{202, 0.00, 1.0}{\strut can}} {\setlength{\fboxsep}{0pt}\colorbox[Hsb]{202, 0.00, 1.0}{\strut I}} {\setlength{\fboxsep}{0pt}\colorbox[Hsb]{202, 0.00, 1.0}{\strut improve}} {\setlength{\fboxsep}{0pt}\colorbox[Hsb]{202, 0.00, 1.0}{\strut my}}\\&
                {\setlength{\fboxsep}{0pt}\colorbox[Hsb]{202, 0.00, 1.0}{\strut English}} {\setlength{\fboxsep}{0pt}\colorbox[Hsb]{202, 0.01, 1.0}{\strut speaking}} {\setlength{\fboxsep}{0pt}\colorbox[Hsb]{202, 0.15, 1.0}{\strut ability}} {\setlength{\fboxsep}{0pt}\colorbox[Hsb]{202, 0.34, 1.0}{\strut ?}}\\ \hline
               {Diversity LSTM}& {\setlength{\fboxsep}{0pt}\colorbox[Hsb]{202, 0.00, 1.0}{\strut How}} {\setlength{\fboxsep}{0pt}\colorbox[Hsb]{202, 0.01, 1.0}{\strut can}} {\setlength{\fboxsep}{0pt}\colorbox[Hsb]{202, 0.02, 1.0}{\strut I}} {\setlength{\fboxsep}{0pt}\colorbox[Hsb]{202, 0.06, 1.0}{\strut improve}} {\setlength{\fboxsep}{0pt}\colorbox[Hsb]{202, 0.08, 1.0}{\strut my}}\\& {\setlength{\fboxsep}{0pt}\colorbox[Hsb]{202, 0.08, 1.0}{\strut English}} {\setlength{\fboxsep}{0pt}\colorbox[Hsb]{202, 0.16, 1.0}{\strut speaking}} {\setlength{\fboxsep}{0pt}\colorbox[Hsb]{202, 0.12, 1.0}{\strut ability}} {\setlength{\fboxsep}{0pt}\colorbox[Hsb]{202, 0.00, 1.0}{\strut ?}} 
            \end{tabular}        \\
        \Xhline{3\arrayrulewidth}
            
         \textbf{Passage}: Sandra went to the garden . Daniel went to the garden.\\
         \textbf{Question}: Where is Sandra?\\
         \textbf{Answer} (Actual \& Predicted): garden\\
         \hline
        {Attention Distribution:}\\
        \begin{tabular}{l|l}
                {Vanilla LSTM}&          {\setlength{\fboxsep}{0pt}\colorbox[Hsb]{202, 0.00, 1.0}{\strut Sandra}} {\setlength{\fboxsep}{0pt}\colorbox[Hsb]{202, 0.00, 1.0}{\strut went}} {\setlength{\fboxsep}{0pt}\colorbox[Hsb]{202, 0.00, 1.0}{\strut to}} {\setlength{\fboxsep}{0pt}\colorbox[Hsb]{202, 0.00, 1.0}{\strut the}} {\setlength{\fboxsep}{0pt}\colorbox[Hsb]{202, 0.00, 1.0}{\strut garden}} {\setlength{\fboxsep}{0pt}\colorbox[Hsb]{202, 0.00, 1.0}{\strut .}}\\& {\setlength{\fboxsep}{0pt}\colorbox[Hsb]{202, 0.00, 1.0}{\strut Daniel}} {\setlength{\fboxsep}{0pt}\colorbox[Hsb]{202, 0.00, 1.0}{\strut went}} {\setlength{\fboxsep}{0pt}\colorbox[Hsb]{202, 0.01, 1.0}{\strut to}} {\setlength{\fboxsep}{0pt}\colorbox[Hsb]{202, 0.02, 1.0}{\strut the}} {\setlength{\fboxsep}{0pt}\colorbox[Hsb]{202, 0.45, 1.0}{\strut garden}} \\ \hline
                {Diversity LSTM}&         {\setlength{\fboxsep}{0pt}\colorbox[Hsb]{202, 0.00, 1.0}{\strut Sandra}} {\setlength{\fboxsep}{0pt}\colorbox[Hsb]{202, 0.00, 1.0}{\strut went}} {\setlength{\fboxsep}{0pt}\colorbox[Hsb]{202, 0.00, 1.0}{\strut to}} {\setlength{\fboxsep}{0pt}\colorbox[Hsb]{202, 0.00, 1.0}{\strut the}} {\setlength{\fboxsep}{0pt}\colorbox[Hsb]{202, 0.46, 1.0}{\strut garden}} {\setlength{\fboxsep}{0pt}\colorbox[Hsb]{202, 0.00, 1.0}{\strut .}}\\& {\setlength{\fboxsep}{0pt}\colorbox[Hsb]{202, 0.00, 1.0}{\strut Daniel}} {\setlength{\fboxsep}{0pt}\colorbox[Hsb]{202, 0.00, 1.0}{\strut went}} {\setlength{\fboxsep}{0pt}\colorbox[Hsb]{202, 0.00, 1.0}{\strut to}} {\setlength{\fboxsep}{0pt}\colorbox[Hsb]{202, 0.00, 1.0}{\strut the}} {\setlength{\fboxsep}{0pt}\colorbox[Hsb]{202, 0.04, 1.0}{\strut garden}}
        \end{tabular}        \\
        \Xhline{3\arrayrulewidth}
    \end{tabular}
    \caption{Samples of Attention distributions from Vanilla and Diversity LSTM models on the Quora Question Paraphrase (QQP) \& Babi 1 datasets.}.
    \label{tab:example}
\end{table}

Attention mechanisms \cite{attn_bahdanau, attention_is_all_you_need} play a very important role in neural network-based models for various Natural Language Processing (NLP) tasks. They not only improve the performance of the model but are also often used to provide insights into the working of a model. Recently, there is a growing debate on whether attention mechanisms can offer \textit{transparency} to a model or not. For example, \startcite{isinterp} and \startcite{notexpl} show that high attention weights need not necessarily correspond to a higher impact on the model's predictions and hence they do not provide a \textit{faithful} explanation for the model's predictions. On the other hand, \startcite{notnotexpl} argues that there is still a possibility that attention distributions may provide a \textit{plausible} explanation for the predictions. In other words, they might provide a \textit{plausible} reconstruction of the model's decision making which can be understood by a human even if it is not faithful to how the model works.

In this work, we begin by analyzing why attention distributions may not faithfully explain the model's predictions. We argue that when the input representations over which an attention distribution is being computed are very similar to each other, the attention weights are not very meaningful. Since the input representations are very similar, even random permutations of the attention weights could lead to similar final context vectors. As a result, the output predictions will not change much even if the attention weights are permuted. We show that this is indeed the case for LSTM based models where the hidden states occupy a narrow cone in the latent space (\textit{i.e.}, the hidden representations are very close to each other). We further observe that for a wide variety of datasets, attention distributions in these models do not even provide a good plausible explanation as they pay significantly high attention to unimportant tokens such as punctuations. This is perhaps due to hidden states capturing a summary of the entire context instead of being specific to their corresponding words.

Based on these observations, we aim to build more transparent and explainable models where the attention distributions provide \textit{faithful} and \textit{plausible} explanations for its predictions. One intuitive way of making the attention distribution more faithful is by ensuring that the hidden representations over which the distribution is being computed are very diverse. Therefore, a random permutation of the attention weights will lead to very different context vectors. To do so, we propose an orthogonalization technique which ensures that the hidden states are farther away from each other in their spatial dimensions. We then propose a more flexible model trained with an additional objective that promotes diversity in the hidden states. Through a series of experiments using $12$ datasets spanning $4$ tasks, we show that our model is more transparent while achieving comparable performance to models containing vanilla LSTM based encoders. Specifically, we show that in our proposed models, attention weights (i) provide useful importance ranking of hidden states (ii) are better indicative of words that are important for the model's prediction (iii) correlate better with gradient-based feature importance methods and (iv) are sensitive to random permutations (as should indeed be the case). 

We further observe that attention weights in our models, in addition to adding transparency to the model, are also more explainable \textit{i.e.} more human-understandable. In Table \ref{tab:example}, we show samples of attention distributions from a Vanilla LSTM and our proposed Diversity LSTM model. We observe that in our models, unimportant tokens such as punctuation marks receive very little attention whereas important words belonging to relevant part-of-speech tags receive greater attention (for example, adjectives in the case of sentiment classification). Human evaluation on the attention from our model shows that humans prefer the attention weights in our Diversity LSTM as providing better explanations than Vanilla LSTM in $72.3\%$, $62.2\%$, $88.4\%$, $99.0\%$ of the samples in Yelp, SNLI, Quora Question Paraphrase and Babi 1 datasets respectively. 
\section{Tasks, Dataset and Models}\label{sec:tasks_datasets}
Our first goal is to understand why existing attention mechanisms with LSTM based encoders fail to provide faithful or plausible explanations for the model's predictions. We experiment on a variety of datasets spanning different tasks; here, we introduce these datasets and tasks and provide a brief recap of the standard LSTM+attention model used for these tasks.
We consider the tasks of \textit{Binary Text classification, Natural Language Inference, Paraphrase Detection, and Question Answering}. We use a total of 12 datasets, most of them being the same as the ones used in \cite{notexpl}. We divide Text classification into Sentiment Analysis and Other Text classification for convenience. 

\textbf{Sentiment Analysis:} We use the \textit{Stanford Sentiment Treebank (SST)} \cite{sst}, \textit{IMDB Movie Reviews} \cite{imdb}, \textit{Yelp} and \textit{Amazon} for sentiment analysis. All these datasets use binary target variable (positive /negative). 

\textbf{Other Text Classification:} We use the \textit{Twitter ADR} \cite{tweet} dataset with 8K tweets where the task is to detect if a tweet describes an adverse drug reaction or not. We use a subset of the \textit{20 Newsgroups} dataset \cite{notexpl} to classify news articles into baseball vs hockey sports categories. From  MIMIC ICD9 \cite{mimic}, we use $2$ datasets: \textit{Anemia}, to determine the type of Anemia (Chronic vs Acute) a patient is diagnosed with and \textit{Diabetes}, to predict whether a patient is diagnosed with Diabetes or not. 

\textbf{Natural Language Inference:} We consider the SNLI dataset \cite{snli} for recognizing textual entailment within sentence pairs. The SNLI dataset has three possible classification labels, \textit{viz} entailment, contradiction and neutral.

\textbf{Paraphrase Detection:} We utilize the Quora Question Paraphrase (QQP) dataset (part of the GLUE benchmark \cite{glue}) with pairs of questions labeled as paraphrased or not. We split the training set into $90:10$ training and validation; and use the original dev set as our test set. 

\textbf{Question Answering:} We made use of all three QA tasks from the \textit{bAbI} dataset \cite{babi}. The tasks consist of answering questions that would require one, two or three supporting statements from the context. The answers are a span in the context. We then use the \textit{CNN News Articles} dataset \cite{cnn} consisting of 90k articles with an average of three questions per article along with their corresponding answers.

\subsection{LSTM Model with Attention}
\label{sec:vanilla_lstm}
Of the above tasks, the text classification tasks require making predictions from a single input sequence (of words) whereas the remaining tasks use pairs of sequences as input. For tasks containing two input sequences, we encode both the sequences $\mathbf{P} = \{w^p_1, \dots, w^p_m\}$ and $\mathbf{Q} = \{w^q_1, \dots, w^q_n\}$ by passing their word embedding through a LSTM encoder \cite{lstm},
\begin{align}
   \nonumber \mathbf{h}^p_t &= \mbox{LSTM}{_{P}}({e(w^p_t)}, \mathbf{h}^p_{t-1})\; \;  \forall t \in [1,m],\\
   \nonumber \mathbf{h}^q_t &= \mbox{LSTM}{_{Q}}({e(w^q_t)}, \mathbf{h}^q_{t-1})\; \;  \forall t \in [1,n],
\end{align}
where ${e(w)}$ represents the word embedding for the word $w$. We attend to the intermediate representations of $\mathbf{P}$, $\mathbf{H}^p = \{\mathbf{h}^p_1, \dots, \mathbf{h}^p_m\}$ $\in \mathbb{R}^{m\times d}$ using the last hidden state $\mathbf{h}^q_n \in \mathbb{R}^{d}$ as the query, using the attention mechanism \cite{attn_bahdanau},
\begin{align}
    \nonumber \tilde{\alpha}_t  &= \mathbf{v}^T \mbox{tanh}(\mathbf{W}_1\mathbf{h}^p_t + \mathbf{W}_2\mathbf{h}^q_n + \mathbf{b}) \; \;  \forall t \in [1,m]\\
    \nonumber \alpha_t &= \mbox{softmax}(\tilde{\alpha}_t) \\
    \nonumber \mathbf{c}_{\alpha} &= \sum_{t=1}^m \alpha_t {\mathbf{h}^p}_t
\end{align}
where $\mathbf{W}_1 \in \mathbb{R}^{d_1 \times d}, \mathbf{W}_2 \in \mathbb{R}^{d_1 \times d}, \mathbf{b} \in \mathbb{R}^{d_1}$ and $\mathbf{v} \in \mathbb{R}^{d_1}$ are learnable parameters. Finally, we use the attended context vector $\mathbf{c}_{\alpha}$ to make a prediction $\hat{y} = \text{softmax} (\mathbf{W}_o \mathbf{c}_{\alpha})$. 

For tasks with a single input sequence, we use a single LSTM to encode the sequence, followed by an attention mechanism (without query) and a final output projection layer. 

\section{Analyzing Attention Mechanisms}
Here, we first investigate the question - \textit{Why Attention distributions may not provide a faithful explanation for the model's predictions?} We later examine whether \textit{Attention distributions can provide {a plausible} explanation for the model's predictions, not necessarily faithful.} 

\subsection{Similarity Measures}
We begin with defining similarity measures in a vector space for ease of analysis. We measure the similarity between a set of vectors $\mathbf{V} = \{\mathbf{v}_1, \dots  ,\mathbf{v}_m\}$ using the \textbf{conicity} measure \cite{conicity1,conicity2} by first computing a vector $\mathbf{v}_i$'s ‘alignment to mean’ (ATM),
\begin{align}
   \nonumber \text{ATM}(\mathbf{v}_i, \mathbf{V}) = \text{cosine}(\mathbf{v}_i, \frac{1}{m} \sum_{j=1}^{m} \mathbf{v}_j)
\end{align}
Conicity is defined as the mean of ATM for all vectors $\mathbf{v}_i \in \mathbf{V}$:
\begin{align}
   \nonumber \text{conicity}(\mathbf{V}) =  \frac{1}{m} \sum_{i=1}^{m} \text{ATM}(\mathbf{v}_i, \mathbf{V})
\end{align}
A high value of conicity indicates that all the vectors are closely aligned with their mean \textit{i.e} they lie in a narrow cone centered at origin.  

\subsection{Attention Mechanisms} \label{sec:importance_ranking}
As mentioned earlier, attention mechanisms learn a weighting distribution over hidden states $\mathbf{H} = \{\mathbf{h}_1,\dots, \mathbf{h}_n\}$ using a scoring function $f$ such as \cite{attn_bahdanau} to obtain an attended context vector $\mathbf{c}_{\alpha}$. 
\begin{align}
    \nonumber \mathbf{c}_{\alpha} = \sum_{t=1}^n \alpha_t {\mathbf{h}}_t; \; \alpha_t = \mbox{softmax}(f(\mathbf{h}_t, \mathbf{h}_{query}))  
\end{align}
The attended context vector is a convex combination of the hidden states which means it will lie within the cone spanned by the hidden states. When the hidden states are highly similar to each other (high conicity), even diverse sets of attention distributions would produce very similar attended context vector $\mathbf{c}_{\alpha}$ as they will always lie within a narrow cone. This could result in outputs  $\hat{y} = \text{softmax} (\mathbf{W}_o \mathbf{c}_{\alpha})$ with very little difference. In other words, when there is a higher conicity in hidden states, the model could produce the same prediction for several diverse sets of attention weights. In such cases, one cannot reliably say that high attention weights on certain input components led the model to its prediction. Later on, in section \ref{sec:importance_ranking}, we show that when using vanilla LSTM encoders where there is higher conicity in hidden states, even when we randomly permute the attention weights, the model output does not change much. 

\begin{figure}
    \centering
    \includegraphics[width=0.3\textwidth]{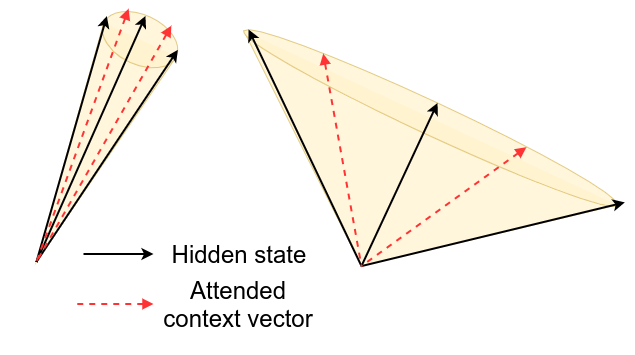}
    \caption{Left: high conicity of hidden states results in similar attended context vectors. Right: low conicity of hidden states results in very different context vectors}
    \label{fig:conicity}
\end{figure}

\subsection{Conicity of LSTMs Hidden States}
We now analyze if the hidden states learned by an LSTM encoder do actually have high conicity. In Table \ref{tab:accuracy_conicity}, we report the average conicity of hidden states learned by an LSTM encoder for various tasks and datasets. For reference, we also compute the average conicity obtained by vectors that are uniformly distributed with respect to direction (isotropic) in the same hidden space. We observe that across all the datasets the hidden states are consistently aligned with each other with conicity values ranging between $0.43$ to $0.77$. In contrast, when there was no dependence between the vectors, the conicity values were much lower with the vectors even being almost orthogonal to its mean in several cases ($\sim 89 \degree$ in Diabetes and Anemia datasets). The existence of high conicity in the learned hidden states of an LSTM encoder is one of the potential reasons why the attention weights in these models are not always \textit{faithful} to its predictions (as even random permutations of the attention weights will result in similar context vectors, $\mathbf{c}_{\alpha}$).  

\subsection{Attention by POS Tags} \label{sec:pos}
We now examine whether attention distributions can provide a plausible explanation for the model's predictions even if it is not faithful. Intuitively, a plausible explanation should ignore unimportant tokens such as punctuation marks and focus on words relevant for the specific task. To examine this, we categorize words in the input sentence by its universal part-of-speech (POS) tag \cite{universalpos} and cumulate attention given to each POS tag over the entire test set. Surprisingly, we find that in several datasets, a significant amount of attention is given to punctuations. On the Yelp, Amazon and QQP datasets, attention mechanisms pay 28.6\%, 34.0\% and 23.0\% of its total attention to punctuations. Notably, punctuations only constitute 11.0\%, 10.5\% and 11.6\% of the total tokens in the respective datasets signifying that learned attention distributions pay substantially greater attention to punctuations than even an uniform distribution. This raises questions on the extent to which attention distributions provide plausible explanations as they attribute model's predictions to tokens that are linguistically insignificant to the context.

One of the potential reasons why the attention distributions are misaligned is that the hidden states might capture a summary of the entire context instead of being specific to their corresponding words as suggested by the high conicity. We later show that attention distributions in our models with low conicity value tend to ignore punctuation marks.  

\section{Orthogonal and Diversity LSTM}
Based on our previous argument that high conicity of hidden states affect the transparency and explainability of attention models, we propose $2$ strategies to obtain reduced similarity in hidden states. 
\begin{figure}
    \centering
    \includegraphics[width=0.4\linewidth]{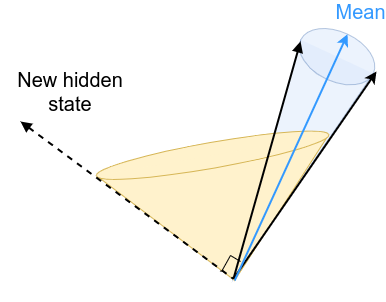}
    \caption{Orthogonal LSTM: Hidden state at a timestep is orthogonal to the mean of previous hidden states}
    \label{fig:ortholstm}
\end{figure}
\subsection{Orthogonalization}
\label{sec:ortho_lstm}
Here, we explicitly ensure low conicity exists between hidden states of an LSTM encoder by orthogonalizing the hidden state at time $t$ with the mean of previous states as illustrated in Figure \ref{fig:ortholstm}. We use the following set of update equations:
\begin{align*}
    \nonumber \mathbf{f}_t &= \sigma (\mathbf{W}_f \mathbf{x}_t + \mathbf{U}_f \mathbf{h}_{t-1} + \mathbf{b}_{f})\\
    \nonumber \mathbf{i}_t &= \sigma (\mathbf{W}_i \mathbf{x}_t + \mathbf{U}_i \mathbf{h}_{t-1} + \mathbf{b}_{i})\\
    \nonumber \mathbf{o}_t &= \sigma (\mathbf{W}_o \mathbf{x}_t + \mathbf{U}_o \mathbf{h}_{t-1} + \mathbf{b}_{o})\\
    \nonumber \hat{\mathbf{c}}_t &= \mbox{tanh} (\mathbf{W}_c \mathbf{x}_t + \mathbf{U}_c \mathbf{h}_{t-1} + \mathbf{b}_{c})\\
    \nonumber \mathbf{c}_t &= \mathbf{f}_{t} \odot \mathbf{c}_{t-1} + \mathbf{i}_t \odot \hat{\mathbf{c}}_t\\
\end{align*}
\begin{align}
    \nonumber \hat{\mathbf{h}}_t &=  \mathbf{o}_t \odot \mbox{tanh}(\mathbf{c}_t)\\
    \overline{\mathbf{h}}_t &= \sum_{i=1}^{t-1} \mathbf{h}_{i}\\
    \label{eq:ortho} \mathbf{h_t} &= \hat{\mathbf{h}}_t - \frac{ \hat{\mathbf{h}}^{T}_t \overline{\mathbf{h}}_t}{\overline{\mathbf{h}}^{T}_t \overline{\mathbf{h}}_t} \overline{\mathbf{h}}_t
\end{align}
where $\mathbf{W}_f, \mathbf{W}_i, \mathbf{W}_o, \mathbf{W}_c \in \mathbb{R}^{d_2 \times d_1}$, $\mathbf{U}_f, \mathbf{U}_i,$ $\mathbf{U}_o, \mathbf{U}_c \in \mathbb{R}^{d_2 \times d_2}$, $\mathbf{b}_f, \mathbf{b}_i, \mathbf{b}_o, \mathbf{b}_c \in \mathbb{R}^{d_2}$, $d_1$ and $d_2$ are the input and hidden dimensions respectively. The key difference from a vanilla LSTM is in the last $2$ equations where we subtract the hidden state vector's $\hat{\mathbf{h}}_t$ component along the mean $\overline{\mathbf{h}}_t$ of the previous states. 

\subsection{Diversity Driven Training}
\label{sec:diversity_training}
The above model imposes a hard orthogonality constraint between the hidden states and the previous states' mean. We also propose a more flexible approach where the model is jointly trained to maximize the log-likelihood of the training data and minimize the conicity of hidden states, 
\begin{align}
    \nonumber L(\theta) =  -p_{model} (y | \mathbf{P}, \mathbf{Q}, \theta) +  \lambda \; \mbox{conicity}(\mathbf{H}^{P})
\end{align}
where $y$ is the ground truth class, $\mathbf{P}$ and $\mathbf{Q}$ are the input sentences, $\mathbf{H}^{P} = \{\mathbf{h}^{p}_{1}, \dots, \mathbf{h}^{p}_{m}\}$ $\in \mathbb{R}^{m\times d}$ contains all the hidden states of the LSTM, $\theta$ is a collection of the model parameters and $p_{model} (.)$ represents the model's output probability. $\lambda$ is a hyperparameter that controls the weight given to diversity in hidden states during training. 


\begin{table*}[bth]
\centering
\resizebox{0.75\textwidth}{!}{
\begin{tabular}{|l|cc|cc|cc|c|c|}
\hline
\multicolumn{1}{|l|}{\multirow{2}{*}{\textbf{Dataset}}} & \multicolumn{2}{c|}{\textbf{LSTM}}                            & \multicolumn{2}{c|}{\textbf{Diversity LSTM}}                  & \multicolumn{2}{c|}{\textbf{Orthogonal LSTM}}                 & \multicolumn{1}{l|}{\textbf{Random}} & \multicolumn{1}{c|}{\textbf{MLP}} \\ \cline{2-9} 
\multicolumn{1}{|l|}{}                                  & \multicolumn{1}{l|}{Accuracy} & \multicolumn{1}{l|}{Conicity} & \multicolumn{1}{l|}{Accuracy} & \multicolumn{1}{l|}{Conicity} & \multicolumn{1}{l|}{Accuracy} & \multicolumn{1}{l|}{Conicity} & \multicolumn{1}{l|}{Conicity}        & \multicolumn{1}{c|}{Accuracy}     \\ \hline
\multicolumn{9}{c}{\textbf{Binary Classification}}                                                                                                                                                                                                                                                                               \\ \hline
SST                                                     & \textbf{81.79}                & 0.68                          & 79.95                         & 0.20                          & 80.05                         & 0.28                          & 0.25                                 & 80.05                             \\
IMDB                                                    & \textbf{89.49}                & 0.69                          & 88.54                         & 0.08                          & 88.71                         & 0.18                          & 0.08                                 & 88.29                             \\
Yelp                                                    & 95.60                         & 0.53                          & 95.40                         & 0.06                          & \textbf{96.00}                & 0.18                          & 0.14                                 & 92.85                             \\
Amazon                                                  & \textbf{93.73}                & 0.50                          & 92.90                         & 0.05                          & 93.04                         & 0.16                          & 0.13                                 & 87.88                             \\
Anemia                                                  & 88.54                         & 0.46                          & 90.09                         & 0.09                          & \textbf{90.17}                & 0.12                          & 0.02                                 & 88.27                             \\
Diabetes                                                & \textbf{92.31}                & 0.61                          & 91.99                         & 0.08                          & 87.05                         & 0.12                          & 0.02                                 & 85.39                             \\
20News                                                  & \textbf{93.55}                & 0.77                          & 91.03                         & 0.15                          & 92.15                         & 0.23                          & 0.13                                 & 87.68                             \\
Tweets                                                  & 87.02                         & 0.77                          & \textbf{87.04}                & 0.24                          & 83.20                         & 0.27                          & 0.24                                 & 80.60                             \\ \hline
\multicolumn{9}{c}{\textbf{Natural Language Inference}}                                                                                                                                                                                                                                                                          \\ \hline
SNLI                                                    & \textbf{78.23}                & 0.56                          & 76.96                         & 0.12                          & 76.46                         & 0.27                          & 0.27                                 & 75.35                             \\ \hline
\multicolumn{9}{c}{\textbf{Paraphrase Detection}}                                                                                                                                                                                                                                                                                \\ \hline
QQP                                                     & \textbf{78.74}                & 0.59                          & 78.40                         & 0.04                          & 78.61                         & 0.33                          & 0.30                                 & 77.78                             \\ \hline
\multicolumn{9}{c}{\textbf{Question Answering}}                                                                                                                                                                                                                                                                                  \\ \hline
bAbI 1                                                  & 99.10                         & 0.56                          & \textbf{100.00}               & 0.07                          & 99.90                         & 0.22                          & 0.19                                 & 42.00                             \\
bAbI 2                                                  & 40.10                         & 0.48                          & 40.20                         & 0.05                          & \textbf{56.10}                & 0.21                          & 0.12                                 & 33.20                             \\
bAbI 3                                                  & 47.70                         & 0.43                          & 50.90                         & 0.10                          & \textbf{51.20}                & 0.12                          & 0.07                                 & 31.60                             \\
CNN                                                     & \textbf{63.07}                & 0.45                          & 58.19                         & 0.06                          & 54.30                         & 0.07                          & 0.04                                 & 37.40   \\ \hline                        
\end{tabular}}
\caption{Accuracy and conicity of Vanilla, Diversity  and Orthogonal LSTM across different datasets. Accuracy of a Multilayered Perceptron (MLP) model and conicity of vectors uniformly distributed with respect to direction is also reported for reference.}
\label{tab:accuracy_conicity}
\end{table*}
\begin{figure}[bth]
    \centering
    \includegraphics[width=0.48\textwidth]{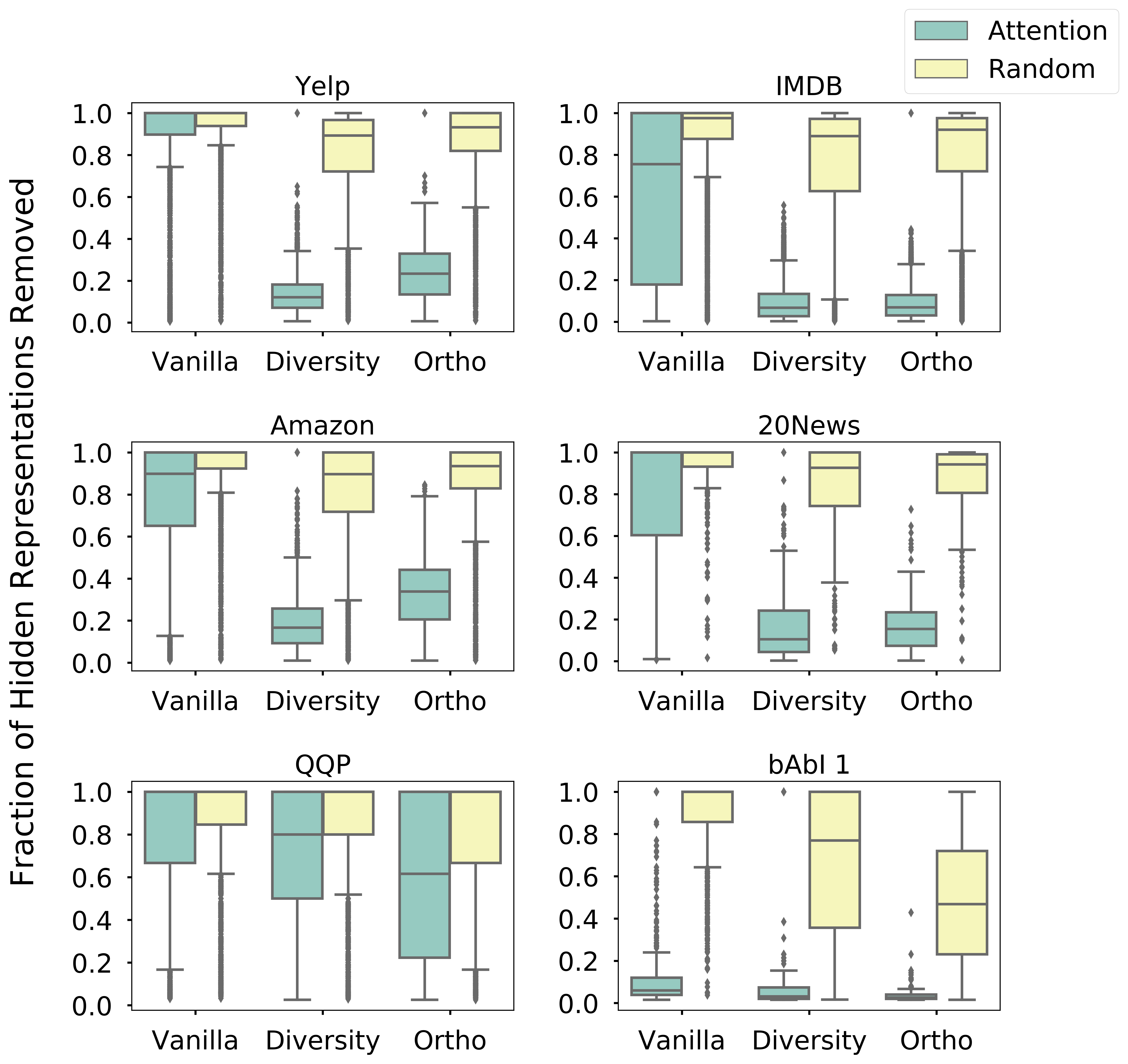}
    \caption{Box plots of fraction of hidden representations removed for a decision flip. Dataset and models are mentioned at the top and bottom of figures. Blue and Yellow indicate the attention and random ranking.}
    \label{fig:importance_ranking}
\end{figure}
\begin{figure}[bth]
    \centering
    \includegraphics[width=0.48\textwidth]{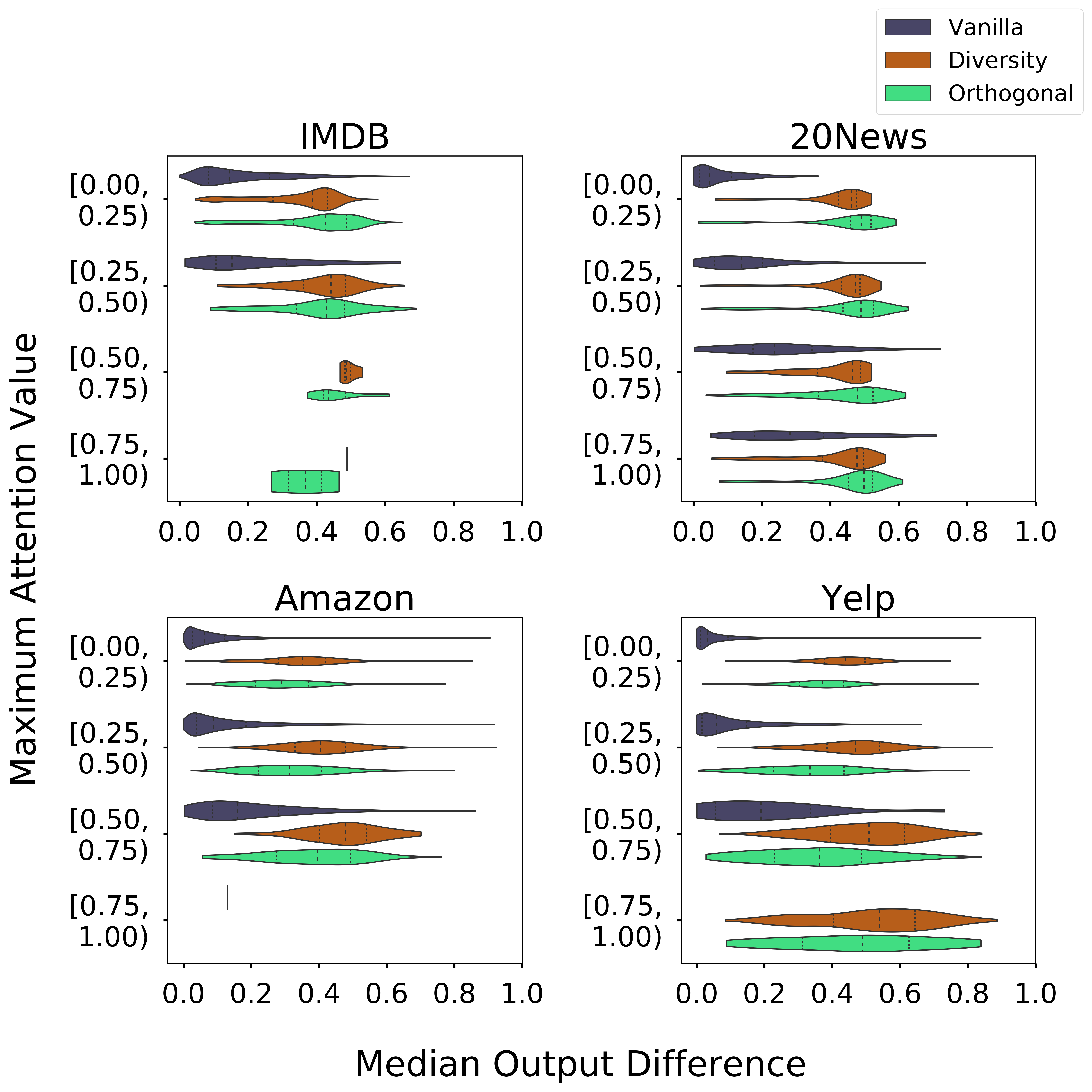}
    \caption{Comparison of Median output difference on randomly permuting the attention weights in the vanilla, Diversity and Orthogonal LSTM models. The Dataset names are mentioned at the top of each figure. Colors indicate the different models as shown legend.}
    \label{fig:permutation}
\end{figure}

\section{Analysis of the model}
We now analyse the proposed models by performing experiments using the tasks and datasets described earlier. Through these experiments we establish that (i) the proposed models perform comparably to vanilla LSTMs (Sec. \ref{sec:empirical_evaluation}) (ii) the attention distributions in the proposed models provide a faithful explanation for the model's predictions (Secs. \ref{sec:importance_ranking} to \ref{sec:gradient_methods}) and (iii) the attention distributions are more explainable and align better with a human's interpretation of the model's prediction (Secs. \ref{sec:pos_tags}, \ref{sec:human_evaluations}). Throughout this section we will compare the following three models:

\noindent\textbf{1. Vanilla LSTM:} The model described in section \ref{sec:vanilla_lstm} which uses the vanilla LSTM.\\
\noindent\textbf{2. Diversity LSTM:} The model described in section \ref{sec:vanilla_lstm} with the vanilla LSTM but trained with the diversity objective described in section \ref{sec:diversity_training}.\\
\noindent\textbf{3. Orthogonal LSTM:}  The model described in section \ref{sec:vanilla_lstm} except that the vanilla LSTM is replaced by the orthogonal LSTM described in section \ref{sec:ortho_lstm}.

\subsection{Implementation Details}
For all datasets except bAbi, we either use pre-trained Glove \cite{glove} or fastText \cite{fasttext} word embeddings with 300 dimensions. For the bAbi dataset, we learn 50 dimensional word embeddings from scratch during training. We use a 1-layered LSTM as the encoder with hidden size of 128 for bAbi and 256 for the other datasets. For the diversity weight $\lambda$, we use a value of 0.1 for SNLI, 0.2 for CNN, and 0.5 for the remaining datasets. We use Adam optimizer with a learning rate of 0.001 and select the best model based on accuracy on the validation split. All the subsequent analysis are performed on the test split. 
\subsection{Empirical evaluation}
\label{sec:empirical_evaluation}
Our main goal is to show that our proposed models provide more faithful and plausible explanations for their predictions. However, before we go there we need to show that the predictive performance of our models is comparable to that of a vanilla LSTM model and significantly better than non-contextual models. In other words, we show that we do not compromise on performance to gain transparency and explainability. We report the performance of our model on the tasks and datasets described in section \ref{sec:tasks_datasets}. In Table \ref{tab:accuracy_conicity}, we report the accuracy and conicity values of vanilla, Diversity and Orthogonal LSTMs on different tasks. We observe that the performance of Diversity LSTM is comparable to that of vanilla LSTM with accuracy values within -7.7\% to +6.7\% (relative) of the vanilla model's accuracy. However, there is a substantial decrease in the conicity values with a drop between 70.6\% to 93.2\% when compared to the vanilla model's conicity. Similarly, for the Orthogonal LSTM, the predictive performance is mostly comparable except for an increase in accuracy by 39.9\% on bAbI 2 and a drop of -13.91\% on CNN. Similar to the Diversity LSTM, the conicity values are much lower than in the vanilla model. We also report the performance of a non-contextual model: Multilayer Perceptron (MLP) + attention in the same table. We observe that both Diversity LSTM and Orthogonal LSTM perform significantly better than the MLP model, especially in difficult tasks such as Question Answering with an average relative increase in accuracy of 73.73\%. Having established that the performance of Diversity and Orthogonal LSTMs is comparable to the vanilla LSTM and significantly better than a Multilayer Perceptron model, we now show that these two models give more faithful explanations for its predictions. 

\subsection{Importance of Hidden Representation}
\label{sec:importance_ranking}

We examine whether attention weights provide a useful importance ranking of hidden representations. We use the intermediate representation erasure by \startcite{isinterp} to evaluate an importance ranking over hidden representations. Specifically, we erase the hidden representations in the descending order of the importance (highest to lowest) until the model's decision changes.
In Figure \ref{fig:importance_ranking}, we report the box plots of the fraction of hidden representations erased for a decision flip when following the ranking provided by attention weights. For reference, we also show the same plots when a random ranking is followed. In several datasets, we observe that a large fraction of the representations have to be erased to obtain a decision flip in the vanilla LSTM model, similar to the observation by
\startcite{isinterp}. This suggests that the hidden representations in the lower end of the attention ranking do play a significant role in the vanilla LSTM model's decision-making process. Hence the usefulness of attention ranking in such models is questionable. In contrast, there is a much quicker decision flip in our Diversity and Orthogonal LSTM models. Thus, in our proposed models, the top elements of the attention ranking are able to concisely describe the model's decisions. This suggests that our attention weights provide a faithful explanation of the model's performance (as higher attention implies higher importance).

In tasks such as paraphrase detection, the model is naturally required to carefully go through the entire sentence to make a decision and thereby resulting in delayed decision flips. In the QA task, the attention ranking in the vanilla LSTM model itself achieves a quick decision flip. On further inspection, we found that this is because these models tend to attend onto answer words which are usually a span in the input passage. So, when the representations corresponding to the answer words are erased, the model can no longer accurately predict the answer resulting in a decision flip. 

Following the work by \cite{notexpl}, we randomly permute the attention weights and observe the difference in the model's output. In Figure \ref{fig:permutation}, we plot the median of Total Variation Distance (TVD) between the output distribution before and after the permutation for different values of maximum attention in the vanilla, Diversity and Orthogonal LSTM models. We observe that randomly permuting the attention weights in the Diversity and Orthogonal LSTM model results in significantly different outputs. However, there is little change in the vanilla LSTM model's output for several datasets suggesting that the attention weights are not so meaningful. The sensitivity of our attention weights to random permutations again suggests that they provide a more faithful explanation for the model's predictions whereas similar outputs raises several questions about the reliability of attention weights in the vanilla LSTM model.

\subsection{Comparison with Rationales}
\begin{table}[]
\scriptsize{
\begin{tabular}{|l|l|l|l|l|}
\hline
\multicolumn{1}{|c|}{\multirow{2}{*}{\textbf{Dataset}}} & \multicolumn{2}{c|}{\textbf{Vanilla LSTM}}                                                                                                                           & \multicolumn{2}{c|}{\textbf{Diversity LSTM}}                                                                                                                         \\ \cline{2-5} 
\multicolumn{1}{|c|}{}                                  & \multicolumn{1}{c|}{\begin{tabular}[c]{@{}c@{}}Rationale\\ Attention\end{tabular}} & \multicolumn{1}{c|}{\begin{tabular}[c]{@{}c@{}}Rationale\\ Length\end{tabular}} & \multicolumn{1}{c|}{\begin{tabular}[c]{@{}c@{}}Rationale\\ Attention\end{tabular}} & \multicolumn{1}{c|}{\begin{tabular}[c]{@{}c@{}}Rationale\\ Length\end{tabular}} \\ \hline
SST                                                     & 0.348                                                                              & 0.240                                                                           & 0.624                                                                              & 0.175                                                                           \\
IMDB                                                    & 0.472                                                                              & 0.217                                                                           & 0.761                                                                              & 0.169                                                                           \\
Yelp                                                    & 0.438                                                                              & 0.173                                                                           & 0.574                                                                              & 0.160                                                                           \\
Amazon                                                  & 0.346                                                                              & 0.162                                                                           & 0.396                                                                              & 0.240                                                                           \\
Anemia                                                  & 0.611                                                                              & 0.192                                                                           & 0.739                                                                              & 0.237                                                                           \\
Diabetes                                                & 0.742                                                                              & 0.458                                                                           & 0.825                                                                              & 0.354                                                                           \\
20News                                                  & 0.627                                                                              & 0.215                                                                           & 0.884                                                                              & 0.173                                                                           \\
Tweets                                                  & 0.284                                                                              & 0.225                                                                           & 0.764                                                                              & 0.306 \\ \hline                                                                         
\end{tabular}}
\caption{Mean Attention given to the generated rationales with their mean lengths (in fraction)}
\label{tab:rationale}
\end{table}

\renewcommand{\arraystretch}{1.0}
\setlength{\tabcolsep}{10pt}
\begin{table*}[]
\centering
\resizebox{0.9\textwidth}{!}{
\begin{tabular}{|l|cc|cc|cc|cc|}
\cline{1-9}
                                                        & \multicolumn{4}{c|}{\textbf{Pearson Correlation $\uparrow$}}                                                                                                                                                                    & \multicolumn{4}{c|}{\textbf{JS Divergence $\downarrow$}}                                                                                                                                                                          \\ \hline
\multicolumn{1}{|c|}{\multirow{2}{*}{\textbf{Dataset}}} & \multicolumn{2}{c|}{\textbf{\begin{tabular}[c]{@{}c@{}}Gradients\\ (Mean $\pm$ Std.)\end{tabular}}} & \multicolumn{2}{c|}{\textbf{\begin{tabular}[c]{@{}c@{}}Integrated Gradients\\ (Mean $\pm$ Std.)\end{tabular}}} & \multicolumn{2}{c|}{\textbf{\begin{tabular}[c]{@{}c@{}}Gradients\\ (Mean $\pm$ Std.)\end{tabular}}} & \multicolumn{2}{c|}{\textbf{\begin{tabular}[c]{@{}c@{}}Integrated Gradients\\ (Mean $\pm$ Std.)\end{tabular}}} \\ \cline{2-9} 
\multicolumn{1}{|c|}{}                                  & Vanilla                                           & Diversity                                       & Vanilla                                                & Diversity                                             & Vanilla                                          & Diversity                                        & Vanilla                                                & Diversity                                             \\ \hline
\multicolumn{9}{c}{\textbf{Text Classification}}                                                                                                                                                                                                                                                                                                                                                                                                                                                    \\ \hline
SST                                                     & 0.71 $\pm$ 0.21                                & 0.83 $\pm$ 0.19                               & 0.62 $\pm$ 0.24                                      & 0.79 $\pm$ 0.22                                     & 0.10 $\pm$ 0.04                                 & 0.08 $\pm$ 0.05                                 & 0.12 $\pm$ 0.05                                       & 0.09 $\pm$ 0.05                                      \\
IMDB                                                    & 0.80 $\pm$ 0.07                                 & 0.89 $\pm$ 0.04                               & 0.68 $\pm$ 0.09                                      & 0.78 $\pm$ 0.07                                     & 0.09 $\pm$ 0.02                                 & 0.09 $\pm$ 0.01                                 & 0.13 $\pm$ 0.02                                       & 0.13 $\pm$ 0.02                                      \\
Yelp                                                    & 0.55 $\pm$ 0.16                                 & 0.79 $\pm$ 0.12                               & 0.40 $\pm$ 0.19                                      & 0.79 $\pm$ 0.14                                     & 0.15 $\pm$ 0.04                                 & 0.13 $\pm$ 0.04                                 & 0.19 $\pm$ 0.05                                       & 0.19 $\pm$ 0.05                                      \\
Amazon                                                  & 0.43 $\pm$ 0.19                                 & 0.77 $\pm$ 0.14                               & 0.43 $\pm$ 0.19                               & 0.77 $\pm$ 0.14                                                                                                              & 0.17 $\pm$ 0.04                                 & 0.12 $\pm$ 0.04                                 & 0.21 $\pm$ 0.06                                       & 0.12 $\pm$ 0.04                                      \\
Anemia                                                  & 0.63 $\pm$ 0.12                                 & 0.72 $\pm$ 0.10                               & 0.43 $\pm$ 0.15                                      & 0.66 $\pm$ 0.11                                     & 0.20 $\pm$ 0.04                                 & 0.19 $\pm$ 0.03                                 & 0.34 $\pm$ 0.05                                       & 0.23 $\pm$ 0.04                                      \\
Diabetes                                                & 0.65 $\pm$ 0.15                                 & 0.76 $\pm$ 0.13                               & 0.55 $\pm$ 0.14                                      & 0.69 $\pm$ 0.18                                     & 0.26 $\pm$ 0.05                                 & 0.20 $\pm$ 0.04                                 & 0.36 $\pm$ 0.04                                       & 0.24 $\pm$ 0.06                                      \\
20News                                                  & 0.72 $\pm$ 0.28                                 & 0.96 $\pm$ 0.08                               & 0.65 $\pm$ 0.32                                      & 0.67 $\pm$ 0.11                                     & 0.15 $\pm$ 0.07                                 & 0.06 $\pm$ 0.04                                 & 0.21 $\pm$ 0.06                                       & 0.07 $\pm$ 0.05                                      \\
Tweets                                                  & 0.65 $\pm$ 0.24                                 & 0.80 $\pm$ 0.21                               & 0.56 $\pm$ 0.25                                      & 0.74 $\pm$ 0.22                                     & 0.08 $\pm$ 0.03                                 & 0.12 $\pm$ 0.07                                 & 0.08 $\pm$ 0.04                                       & 0.15 $\pm$ 0.06                                      \\ \hline
\multicolumn{9}{c}{\textbf{Natural Language Inference}}                                                                                                                                                                                                                                                                                                                                                                                                                                             \\ \hline
SNLI                                                    & 0.58 $\pm$ 0.33                                 & 0.51 $\pm$ 0.35                               & 0.38 $\pm$ 0.40                                      & 0.26 $\pm$ 0.39                                     & 0.11 $\pm$ 0.07                                 & 0.10 $\pm$ 0.06                                 & 0.16 $\pm$ 0.09                                       & 0.13 $\pm$ 0.06                                      \\ \hline
\multicolumn{9}{c}{\textbf{Paraphrase Detection}}                                                                                                                                                                                                                                                                                                                                                                                                                                                   \\ \hline
QQP                                                     & 0.19 $\pm$ 0.34                                 & 0.58 $\pm$ 0.31                               & -0.06 $\pm$ 0.34                                     & 0.21 $\pm$ 0.36                                     & 0.15 $\pm$ 0.08                                 & 0.10 $\pm$ 0.05                                 & 0.19 $\pm$ 0.10                                       & 0.15 $\pm$ 0.06                                      \\ \hline
\multicolumn{9}{c}{\textbf{Question Answering}}                                                                                                                                                                                                                                                                                                                                                                                                                                                     \\ \hline
Babi 1                                                  & 0.56 $\pm$ 0.34                                 & 0.91 $\pm$ 0.10                               & 0.33 $\pm$ 0.37                                      & 0.91 $\pm$ 0.10                                     & 0.33 $\pm$ 0.12                                 & 0.21 $\pm$ 0.08                                 & 0.43 $\pm$ 0.13                                       & 0.24 $\pm$ 0.08                                      \\
Babi 2                                                  & 0.16 $\pm$ 0.23                                 & 0.70 $\pm$ 0.13                               & 0.05 $\pm$ 0.22                                      & 0.75 $\pm$ 0.10                                     & 0.53 $\pm$ 0.09                                 & 0.23 $\pm$ 0.06                                 & 0.58 $\pm$ 0.09                                       & 0.19 $\pm$ 0.05                                      \\
Babi 3                                                  & 0.39 $\pm$ 0.24                                 & 0.67 $\pm$ 0.19                               & -0.01 $\pm$ 0.08                                     & 0.47 $\pm$ 0.25                                     & 0.46 $\pm$ 0.08                                 & 0.37 $\pm$ 0.07                                 & 0.64 $\pm$ 0.05                                       & 0.41 $\pm$ 0.08                                      \\
CNN                                                     & 0.58 $\pm$ 0.25                                 & 0.75 $\pm$ 0.20                               & 0.45 $\pm$ 0.28                                      & 0.66 $\pm$ 0.23                                     & 0.22 $\pm$ 0.07                                 & 0.17 $\pm$ 0.08                                 & 0.30 $\pm$ 0.10                                       & 0.21 $\pm$ 0.10    \\ \hline                                
\end{tabular}}
\caption{Mean and standard deviation of Pearson correlation and Jensen–Shannon divergence between Attention weights and Gradients/Integrated Gradients in Vanilla and Diversity LSTM models}
\label{tab:correlations}
\end{table*}
For tasks with a single input sentence, we analyze how much attention is given to words in the sentence that are important for the prediction. Specifically, we select a minimum subset of words in the input sentence with which the model can accurately make predictions. We then compute the total attention that is paid to these words. These set of words, also known as rationales, are obtained from an extractive rationale generator \cite{rationale} that is trained using the REINFORCE algorithm \cite{reinforce} to maximize the following reward:
\begin{align}
    \nonumber R = p_{model} (y | \mathbf{Z}) - \alpha  ||\mathbf{Z}||
\end{align}
where $y$ is the ground truth class, $\mathbf{Z}$ is the extracted rationale, $||\mathbf{Z}||$ represents the length of the rationale, $p_{model} (.)$ represents the classification model's output probability, $\alpha$ is a hyperparameter that penalizes long rationales. With a fixed $\alpha$, we trained generators to extract rationales from the vanilla and Diversity LSTM models. We observed that the accuracy of predictions made from the extracted rationales was within 5\% of the accuracy made from the entire sentences. In Table \ref{tab:rationale}, we report the mean length (in fraction) of the rationales and the mean attention given to them in the vanilla and Diversity LSTM models. In general, we observe that the Diversity LSTM model provides much higher attention to rationales which are even often shorter than the vanilla LSTM model's rationales. On average, the Diversity LSTM model provides 53.52 \% (relative) more attention to rationales than the vanilla LSTM across the 8 Text classification datasets. Thus, the attention weights in the Diversity LSTM are able to better indicate words that are important for making predictions. 


\subsection{Comparison with attribution methods}
\label{sec:gradient_methods}
We now examine how well our attention weights agree with attribution methods such as gradients and integrated gradients \cite{ig}. For every input word, we compute these attributions and normalize them to obtain a distribution over the input words. We then compute the Pearson correlation and JS divergence between the attribution distribution and the attention distribution. We note that Kendall $\tau$ as used by \cite{notexpl} often results in misleading correlations because the ranking at the tail end of the distributions contributes to a significant noise. In Table \ref{tab:correlations}, we report the mean and standard deviation of these Pearson correlations and JS divergence in the vanilla and Diversity LSTMs across different datasets. We observe that attention weights in Diversity LSTM better agree with gradients with an average (relative) 64.84\% increase in Pearson correlation and an average (relative) 17.18\% decrease in JS divergence over the vanilla LSTM across the datasets. Similar trends follow for Integrated Gradients.  

\subsection{Analysis by POS tags}
\label{sec:pos_tags}
\begin{figure}
    \centering
    \includegraphics[width=1.0\linewidth]{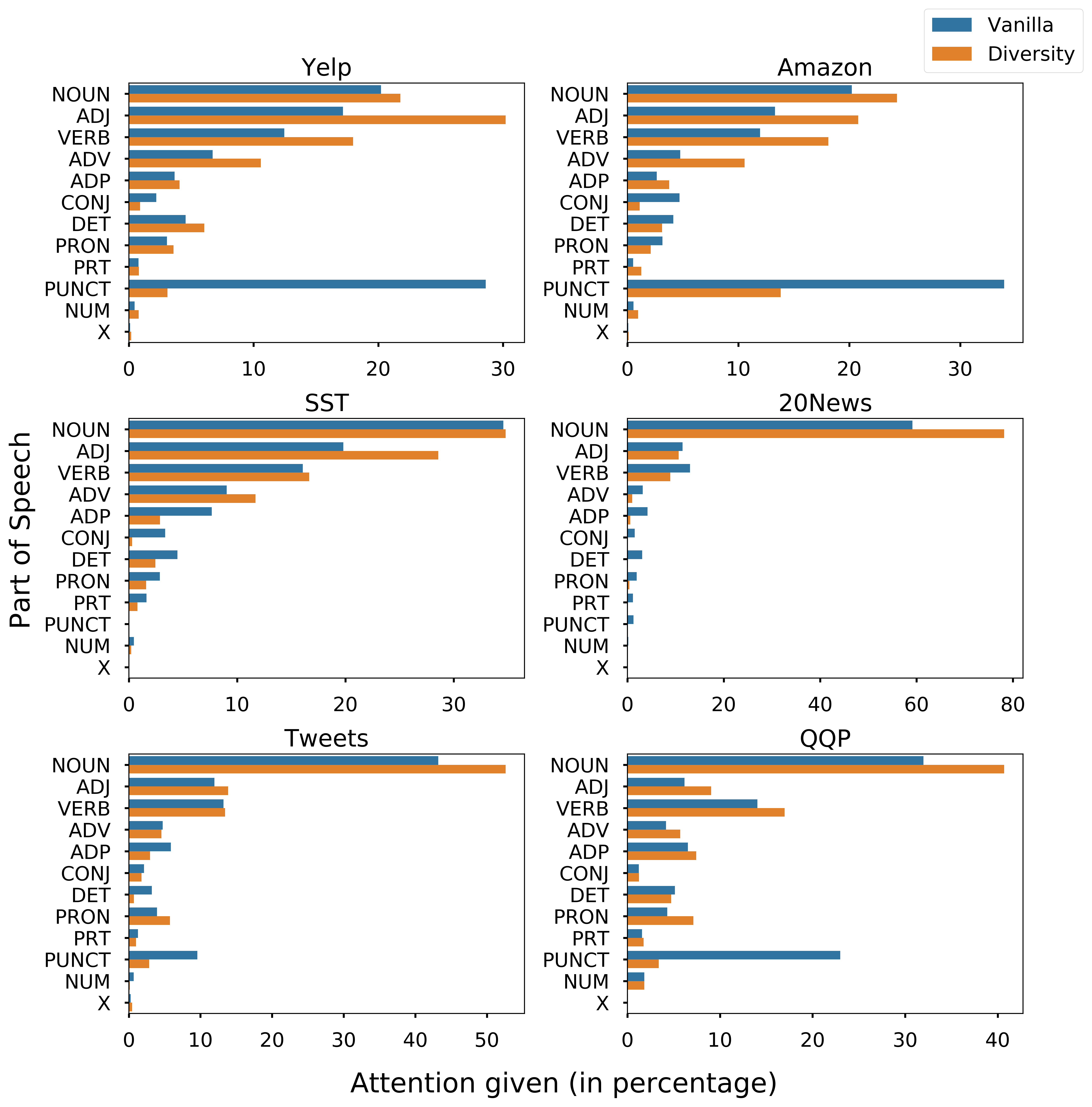}
    \caption{Distribution of cumulative attention given to different part-of-speech tags in the test dataset. Blue and Orange indicate the vanilla and Diversity LSTMs.}
    \label{fig:pos}
\end{figure}

Figure \ref{fig:pos} shows the distribution of attention given to different POS tags across different datasets. We observe that the attention given to punctuation marks is significantly reduced from 28.6\%, 34.0\% and 23.0\% in the vanilla LSTM to 3.1\%, 13.8\% and 3.4\% in the Diversity LSTM on the Yelp, Amazon and QQP datasets respectively. In the sentiment classification task, Diversity LSTM pays greater attention to the adjectives, which usually play a crucial role in deciding the polarity of a sentence. Across the four sentiment analysis datasets, Diversity LSTM gives an average of 49.27 \% (relative) more attention to adjectives than the vanilla LSTM. Similarly, for the other text classification tasks where nouns play an important role, we observe higher attention to nouns. 

\begin{table}[]
\resizebox{0.48\textwidth}{!}{\begin{tabular}{|l|l|l|l|}
\hline
\multicolumn{1}{|c|}{\multirow{2}{*}{\textbf{Dataset}}} & \multicolumn{1}{c|}{\textbf{Overall}} & \multicolumn{1}{c|}{\textbf{Completness}} & \multicolumn{1}{c|}{\textbf{Correctness}} \\ \cline{2-4} 
\multicolumn{1}{|c|}{}                                  & \multicolumn{1}{c|}{Vanilla/Divers.}  & Vanilla/Divers.                           & Vanilla/Divers.                           \\ \hline
Yelp                                                    & 27.7\% / 72.3\%                       & 35.1\% / 64.9\%                           & 10.5\% / 89.5\%                           \\
SNLI                                                    & 37.8\% / 62.2\%                       & 32.3\% / 67.7\%                           & 38.9\% / 61.1\%                           \\
QQP                                                     & 11.6\% / 88.4\%                       & 11.8\% / 88.2\%                           & 7.9\% / 92.1\%                            \\
bAbI 1                                                  & 1.0\% / 99.0\%                        & 4.2\% / 95.8\%                            & 1.0\% / 99.0\%  \\ \hline                         
\end{tabular}}
\caption{Percentage preference given to Vanilla vs Diversity model by human annotators based on 3 criteria}
\label{tab:humaneval}
\end{table}

\subsection{Human Evaluations}
\label{sec:human_evaluations}
We conducted human evaluations to compare the extent to which attention distributions from the vanilla and Diversity LSTMs provide plausible explanations. We randomly sampled 200 data points each from the test sets of Yelp, SNLI, QQP, and bAbI1. Annotators were shown the input sentence, the attention heatmaps, and predictions made by the vanilla and Diversity LSTMs and were asked to choose the attention heatmap that better explained the model's prediction on 3 criteria 1) Overall - which heatmap is better in explaining the prediction overall 2) Completeness - which heatmap highlights all the words necessary for the prediction. 3) Correctness - which heatmap only highlights the important words and not unnecessary words. Annotators were given the choice to skip a sample in case they were unable to make a clear decision. A total of 15 in-house annotators participated in the human evaluation study. The annotators were Computer Science graduates competent in English. We had $3$ annotators for each sample and the final decision was taken based on majority voting. In Table \ref{tab:humaneval}, we report the percentage preference given to the vanilla and Diversity LSTM models on the Yelp, SNLI, QQP, and bAbI 1 datasets; the attention distributions from Diversity LSTM significantly outperforms the attention from vanilla LSTM across all the datasets and criteria.

 \section{Related work}

Our work in many ways can be seen as a continuation to the recent studies \cite{isinterp, notexpl, notnotexpl} on the subject of interpretability of attention. Several other works \cite{sparse1,sparse2,sparse3,sparse4,sparse5, sparse6} focus on improving the interpretability of attention distributions by inducing sparsity. However, the extent to which sparse attention distributions actually offer faithful and plausible explanations haven't been studied in detail. Few works \cite{rationaleattn} map attention distributions to human annotated rationales. Our work on the other hand does not require any additional supervision. Work by \cite{interpretablelstm} focus on developing interpretable LSTMs specifically for multivariate time series analysis. Several other works \cite{whatDoesBertlook,Transformerattention,BERTRT,are16betterthan1,whatbertlearns,tranfformersattnkernal} analyze attention distributions and attention heads learned by transformer language models. The idea of orthogonalizing representations in
an LSTM have been used by \cite{Nema2017DiversityDA} but they use a different diversity model in the context of improving performance of
Natural Language Generation models

\section{Conclusion \& Future work}
In this work, we have analyzed why existing attention distributions can neither provide a faithful nor a plausible explanation for the model's predictions.  We showed that hidden representations learned by LSTM encoders tend to be highly similar across different timesteps, thereby affecting the interpretability of attention weights. We proposed two techniques to effectively overcome this shortcoming and showed that attention distributions in the resulting models provide more faithful and plausible explanations. As future work, we would like to extend our analysis and proposed techniques to more complex models and downstream tasks. 

\section*{Acknowledgements}
 We would like to thank Department of Computer Science and Engineering, IIT Madras and Robert Bosch Center for Data Sciences and Artificial Intelligence, IIT Madras (RBC-DSAI) for providing us sufficient resources. We acknowledge Google for supporting Preksha Nema's contribution through their Google India Ph.D. fellowship program. We also express our gratitude to the annotators who participated in human evaluations.

\bibliography{acl2020}

\begin{thebibliography}{37}
\expandafter\ifx\csname natexlab\endcsname\relax\def\natexlab#1{#1}\fi

\bibitem[{Bahdanau et~al.(2014)Bahdanau, Cho, and Bengio}]{attn_bahdanau}
Dzmitry Bahdanau, Kyunghyun Cho, and Yoshua Bengio. 2014.
\newblock Neural machine translation by jointly learning to align and
  translate.
\newblock \emph{CoRR}, abs/1409.0473.

\bibitem[{Bao et~al.(2018)Bao, Chang, Yu, and Barzilay}]{rationaleattn}
Yujia Bao, Shiyu Chang, Mo~Yu, and Regina Barzilay. 2018.
\newblock Deriving machine attention from human rationales.
\newblock In \emph{EMNLP}.

\bibitem[{Bowman et~al.(2015)Bowman, Angeli, Potts, and Manning}]{snli}
Samuel~R. Bowman, Gabor Angeli, Christopher Potts, and Christopher~D. Manning.
  2015.
\newblock A large annotated corpus for learning natural language inference.
\newblock In \emph{EMNLP}.

\bibitem[{Chandrahas et~al.(2018)Chandrahas, Sharma, and Talukdar}]{conicity1}
Chandrahas, Aditya Sharma, and Partha~P. Talukdar. 2018.
\newblock Towards understanding the geometry of knowledge graph embeddings.
\newblock In \emph{ACL}.

\bibitem[{Clark et~al.(2019)Clark, Khandelwal, Levy, and
  Manning}]{whatDoesBertlook}
Kevin Clark, Urvashi Khandelwal, Omer Levy, and Christopher~D. Manning. 2019.
\newblock What does bert look at? an analysis of bert's attention.
\newblock \emph{ArXiv}, abs/1906.04341.

\bibitem[{Guo et~al.(2019)Guo, Lin, and Antulov-Fantulin}]{interpretablelstm}
Tian Guo, Tao Lin, and Nino Antulov-Fantulin. 2019.
\newblock Exploring interpretable lstm neural networks over multi-variable
  data.
\newblock In \emph{ICML}.

\bibitem[{Hermann et~al.(2015)Hermann, Kocisk{\'y}, Grefenstette, Espeholt,
  Kay, Suleyman, and Blunsom}]{cnn}
Karl~Moritz Hermann, Tom{\'a}s Kocisk{\'y}, Edward Grefenstette, Lasse
  Espeholt, Will Kay, Mustafa Suleyman, and Phil Blunsom. 2015.
\newblock Teaching machines to read and comprehend.
\newblock In \emph{NIPS}.

\bibitem[{Hochreiter and Schmidhuber(1997)}]{lstm}
Sepp Hochreiter and J{\"u}rgen Schmidhuber. 1997.
\newblock Long short-term memory.
\newblock \emph{Neural Computation}, 9:1735--1780.

\bibitem[{Jain and Wallace(2019)}]{notexpl}
Sarthak Jain and Byron~C. Wallace. 2019.
\newblock Attention is not explanation.
\newblock In \emph{NAACL-HLT}.

\bibitem[{Jawahar et~al.(2019)Jawahar, Sagot, and Seddah}]{whatbertlearns}
Ganesh Jawahar, Beno{\^i}t Sagot, and Djam{\'e} Seddah. 2019.
\newblock What does bert learn about the structure of language?
\newblock In \emph{ACL}.

\bibitem[{Johnson et~al.(2016)Johnson, Pollard, Shen, wei H.~Lehman, Feng,
  Ghassemi, Moody, Szolovits, Celi, and Mark}]{mimic}
Alistair E.~W. Johnson, Tom~J. Pollard, Lu~Shen, Li~wei H.~Lehman, Mengling
  Feng, Mohammad~M. Ghassemi, Benjamin Moody, Peter Szolovits, Leo~Anthony
  Celi, and Roger~G. Mark. 2016.
\newblock Mimic-iii, a freely accessible critical care database.
\newblock In \emph{Scientific data}.

\bibitem[{Lei et~al.(2016)Lei, Barzilay, and Jaakkola}]{rationale}
Tao Lei, Regina Barzilay, and Tommi~S. Jaakkola. 2016.
\newblock Rationalizing neural predictions.
\newblock In \emph{EMNLP}.

\bibitem[{Maas et~al.(2011)Maas, Daly, Pham, Huang, Ng, and Potts}]{imdb}
Andrew~L. Maas, Raymond~E. Daly, Peter~T. Pham, Dan Huang, Andrew~Y. Ng, and
  Christopher Potts. 2011.
\newblock Learning word vectors for sentiment analysis.
\newblock In \emph{ACL}.

\bibitem[{Malaviya et~al.(2018)Malaviya, Ferreira, and Martins}]{sparse3}
Chaitanya Malaviya, Pedro Ferreira, and Andr{\'e} F.~T. Martins. 2018.
\newblock Sparse and constrained attention for neural machine translation.
\newblock In \emph{ACL}.

\bibitem[{Martins and Astudillo(2016)}]{sparse2}
Andr{\'e} F.~T. Martins and Ram{\'o}n~Fern{\'a}ndez Astudillo. 2016.
\newblock From softmax to sparsemax: A sparse model of attention and
  multi-label classification.
\newblock \emph{ArXiv}, abs/1602.02068.

\bibitem[{Maruf et~al.(2019)Maruf, Martins, and Haffari}]{sparse5}
Sameen Maruf, Andr{\'e} F.~T. Martins, and Gholamreza Haffari. 2019.
\newblock Selective attention for context-aware neural machine translation.
\newblock In \emph{NAACL-HLT}.

\bibitem[{Michel et~al.(2019)Michel, Levy, and Neubig}]{are16betterthan1}
Paul Michel, Omer Levy, and Graham Neubig. 2019.
\newblock Are sixteen heads really better than one?
\newblock \emph{ArXiv}, abs/1905.10650.

\bibitem[{Mikolov et~al.(2018)Mikolov, Grave, Bojanowski, Puhrsch, and
  Joulin}]{fasttext}
Tomas Mikolov, Edouard Grave, Piotr Bojanowski, Christian Puhrsch, and Armand
  Joulin. 2018.
\newblock Advances in pre-training distributed word representations.
\newblock In \emph{Proceedings of the International Conference on Language
  Resources and Evaluation (LREC 2018)}.

\bibitem[{Nema et~al.(2017)Nema, Khapra, Laha, and
  Ravindran}]{Nema2017DiversityDA}
Preksha Nema, Mitesh~M. Khapra, Anirban Laha, and Balaraman Ravindran. 2017.
\newblock Diversity driven attention model for query-based abstractive
  summarization.
\newblock In \emph{ACL}.

\bibitem[{Niculae and Blondel(2017)}]{sparse4}
Vlad Niculae and Mathieu Blondel. 2017.
\newblock A regularized framework for sparse and structured neural attention.
\newblock In \emph{NIPS}.

\bibitem[{Nikfarjam et~al.(2015)Nikfarjam, Sarker, O'Connor, Ginn, and
  Gonzalez-Hernandez}]{tweet}
Azadeh Nikfarjam, Abeed Sarker, Karen O'Connor, Rachel~E. Ginn, and Graciela
  Gonzalez-Hernandez. 2015.
\newblock Pharmacovigilance from social media: mining adverse drug reaction
  mentions using sequence labeling with word embedding cluster features.
\newblock In \emph{JAMIA}.

\bibitem[{Pennington et~al.(2014)Pennington, Socher, and Manning}]{glove}
Jeffrey Pennington, Richard Socher, and Christopher~D. Manning. 2014.
\newblock Glove: Global vectors for word representation.
\newblock In \emph{Empirical Methods in Natural Language Processing (EMNLP)},
  pages 1532--1543.

\bibitem[{Peters et~al.(2018)Peters, Niculae, and Martins}]{sparse6}
Ben Peters, Vlad Niculae, and Andr{\'e} F.~T. Martins. 2018.
\newblock Interpretable structure induction via sparse attention.
\newblock In \emph{BlackboxNLP@EMNLP}.

\bibitem[{Petrov et~al.(2011)Petrov, Das, and McDonald}]{universalpos}
Slav Petrov, Dipanjan Das, and Ryan~T. McDonald. 2011.
\newblock A universal part-of-speech tagset.
\newblock In \emph{LREC}.

\bibitem[{Sai et~al.(2019)Sai, Gupta, Khapra, and Srinivasan}]{conicity2}
Ananya Sai, Mithun~Das Gupta, Mitesh~M. Khapra, and Mukundhan Srinivasan. 2019.
\newblock Re-evaluating adem: A deeper look at scoring dialogue responses.
\newblock In \emph{AAAI}.

\bibitem[{Serrano and Smith(2019)}]{isinterp}
Sofia Serrano and Noah~A. Smith. 2019.
\newblock Is attention interpretable?
\newblock In \emph{ACL}.

\bibitem[{Shao et~al.(2019)Shao, Meng, Li, Zhang, Li, Wang, and Luo}]{sparse1}
Wenqi Shao, Tianjian Meng, Jingyu Li, Ruimao Zhang, Yudian Li, Xiaogang Wang,
  and Ping Luo. 2019.
\newblock Ssn: Learning sparse switchable normalization via sparsestmax.
\newblock In \emph{CVPR}.

\bibitem[{Socher et~al.(2013)Socher, Perelygin, Wu, Chuang, Manning, Ng, and
  Potts}]{sst}
Richard Socher, Alex Perelygin, Jean Wu, Jason Chuang, Christopher~D. Manning,
  Andrew~Y. Ng, and Christopher Potts. 2013.
\newblock Recursive deep models for semantic compositionality over a sentiment
  treebank.
\newblock In \emph{EMNLP}.

\bibitem[{Sundararajan et~al.(2017)Sundararajan, Taly, and Yan}]{ig}
Mukund Sundararajan, Ankur Taly, and Qiqi Yan. 2017.
\newblock Axiomatic attribution for deep networks.
\newblock In \emph{ICML}.

\bibitem[{Sutton et~al.(1999)Sutton, McAllester, Singh, and
  Mansour}]{reinforce}
Richard~S. Sutton, David~A. McAllester, Satinder~P. Singh, and Yishay Mansour.
  1999.
\newblock Policy gradient methods for reinforcement learning with function
  approximation.
\newblock In \emph{NIPS}.

\bibitem[{Tenney et~al.(2019)Tenney, Das, and Pavlick}]{BERTRT}
Ian Tenney, Dipanjan Das, and Ellie Pavlick. 2019.
\newblock Bert rediscovers the classical nlp pipeline.
\newblock In \emph{ACL}.

\bibitem[{Tsai et~al.(2019)Tsai, Bai, Yamada, Morency, and
  Salakhutdinov}]{tranfformersattnkernal}
Yao-Hung Tsai, Shaojie Bai, Makoto Yamada, Louis-Philippe Morency, and Ruslan
  Salakhutdinov. 2019.
\newblock Empirical study of transformer’s attention mechanism via the lens
  of kernel.
\newblock In \emph{IJCNLP 2019}.

\bibitem[{Vaswani et~al.(2017)Vaswani, Shazeer, Parmar, Uszkoreit, Jones,
  Gomez, Kaiser, and Polosukhin}]{attention_is_all_you_need}
Ashish Vaswani, Noam Shazeer, Niki Parmar, Jakob Uszkoreit, Llion Jones,
  Aidan~N. Gomez, Lukasz Kaiser, and Illia Polosukhin. 2017.
\newblock Attention is all you need.
\newblock In \emph{NIPS}.

\bibitem[{Vig and Belinkov(2019)}]{Transformerattention}
Jesse Vig and Yonatan Belinkov. 2019.
\newblock Analyzing the structure of attention in a transformer language model.
\newblock \emph{ArXiv}, abs/1906.04284.

\bibitem[{Wang et~al.(2018)Wang, Singh, Michael, Hill, Levy, and Bowman}]{glue}
Alex Wang, Amanpreet Singh, Julian Michael, Felix Hill, Omer Levy, and
  Samuel~R. Bowman. 2018.
\newblock Glue: A multi-task benchmark and analysis platform for natural
  language understanding.
\newblock In \emph{BlackboxNLP@EMNLP}.

\bibitem[{Weston et~al.(2015)Weston, Bordes, Chopra, and Mikolov}]{babi}
Jason Weston, Antoine Bordes, Sumit Chopra, and Tomas Mikolov. 2015.
\newblock Towards ai-complete question answering: A set of prerequisite toy
  tasks.
\newblock \emph{CoRR}, abs/1502.05698.

\bibitem[{Wiegreffe and Pinter(2019)}]{notnotexpl}
Sarah Wiegreffe and Yuval Pinter. 2019.
\newblock Attention is not not explanation.
\newblock \emph{ArXiv}, abs/1908.04626.

\end{thebibliography}
\bibliographystyle{acl_natbib}
\end{document}